\documentclass{article}

\usepackage{arxiv}

\usepackage{graphicx}
\usepackage[utf8]{inputenc} % allow utf-8 input
\usepackage[T1]{fontenc}    % use 8-bit T1 fonts
\usepackage[colorlinks,allcolors=blue]{hyperref}       % hyperlinks
\usepackage{url}            % simple URL typesetting
\usepackage{booktabs}       % professional-quality tables
\usepackage{amsfonts}       % blackboard math symbols
\usepackage{nicefrac}       % compact symbols for 1/2, etc.
\usepackage{microtype}      % microtypography
\usepackage{lipsum}		% Can be removed after putting your text content
\usepackage{graphicx}
\usepackage{natbib}
\usepackage{doi}
\usepackage{tabularx,booktabs}
\usepackage{algorithm}
\usepackage{algpseudocode} %not in acm list
\usepackage{multirow}
\usepackage{xcolor}
\usepackage{multicol}
\usepackage{subcaption}
\usepackage{tikz} % not in acm list
\usepackage{amsmath}
\usepackage{textcomp}
\usepackage{libertine}
\usepackage{comment}
\usepackage{amssymb}
\algnewcommand\algorithmicforeach{\textbf{for each}}
\algnewcommand\algorithmicdoparallel{\textbf{do in parallel}}
\algdef{S}[FOR]{ForEach}[1]{\algorithmicforeach\ #1\ \algorithmicdo}
\algdef{S}[FOR]{ForEachParallel}[1]{\algorithmicforeach\ #1\ \algorithmicdoparallel}
\algnewcommand{\sIf}[2]{%\sIf{<if>}{<then>}
  \State \algorithmicif\ #1\ \algorithmicthen\ #2}
\newcommand{\assign}{\ensuremath{\leftarrow}}

\algnewcommand{\sIfElse}[3]{%\sIfElse{<if>}{<then>}{<else>}
  \State \algorithmicif\ #1\ \algorithmicthen\ #2 \ \algorithmicelse\ #3}
\graphicspath{graphics}
\usepackage{xspace}

\title{A Study of Scalarisation Techniques for Multi-Objective QUBO Solving\thanks{Please cite proceedings for OR2022 (\url{https://www.or2022.de/)} }}

\author{ \href{https://orcid.org/0000-0003-0854-4777}{\includegraphics[scale=0.06]{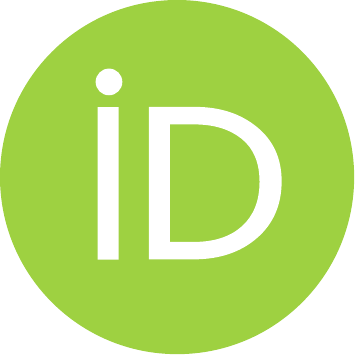}\hspace{1mm}Mayowa Ayodele}\\
	Fujitsu Research of Europe Ltd.\\
	Slough\\
	United Kingdom\\
	\texttt{mayowa.ayodele@fujitsu.com} \\
	\And
	\href{https://orcid.org/0000-0003-1236-3143}{\includegraphics[scale=0.06]{orcid.pdf}\hspace{1mm}Richard Allmendinger} \\
	The University of Manchester\\
	Manchester\\
	United Kingdom \\
	\texttt{richard.allmendinger@manchester.ac.uk} \\
		\And
	\href{https://orcid.org/0000-0001-9974-1295}{\includegraphics[scale=0.06]{orcid.pdf}\hspace{1mm}Manuel López-Ibáñez} \\
	The University of Manchester\\
	Manchester\\
	United Kingdom \\
	\texttt{manuel.lopez-ibanez@manchester.ac.uk} \\
		\And
	\href{https://orcid.org/0000-0002-5777-7756}{\includegraphics[scale=0.06]{orcid.pdf}\hspace{1mm}Matthieu Parizy} \\
	Fujitsu Ltd.\\
	Kawasaki\\
	Japan \\
	\texttt{parizy.matthieu@fujitsu.com} \\
}

\renewcommand{\shorttitle}{A Study of Scalarisation Techniques for Multi-Objective QUBO Solving}
\begin{document}
\maketitle

\begin{abstract}
In recent years, there has been significant research interest in solving Quadratic Unconstrained Binary Optimisation (QUBO) problems. Physics-inspired optimisation algorithms have been proposed for deriving optimal or sub-optimal solutions to QUBOs. These methods are particularly attractive within the context of using specialised hardware, such as quantum computers, application specific CMOS and other high performance computing resources for solving optimisation problems. Examples of such solvers are D-wave’s Quantum Annealer and Fujitsu’s Digital Annealer. These solvers are then applied to QUBO formulations of combinatorial optimisation problems. Quantum and quantum-inspired optimisation algorithms have shown promising performance when applied to academic benchmarks as well as real-world problems. However, QUBO solvers are single objective solvers. To make them more efficient at solving problems with multiple objectives, a decision on how to convert such multi-objective problems to single-objective problems need to be made. In this study, we compare methods of deriving scalarisation weights when combining two objectives of the cardinality constrained mean-variance portfolio optimisation problem into one. We show significant performance improvement (measured in terms of hypervolume) when using a method that iteratively fills the largest space in the Pareto front compared to a na\"{i}ve approach using uniformly generated weights.
\end{abstract}



\section{Introduction}

In recent years, there has been significant research interest in solving Quadratic Unconstrained Binary Optimisation (QUBO) formulations of optimisation problems. This is a common formulation used by hardware solvers classified as quantum or quantum-inspired machines. They have been shown to achieve a speed up compared to classical optimisation algorithms implemented on general purpose computers \cite{ayodele2022comparing}. Ising machines such as Fujitsu's Digital Annealer (DA)~\cite{da3} and D-wave's Quantum Annealer \cite{mcgeoch2020d} are single objective solvers. Many optimisation problems however have more than one objective, e.g. the Cardinality Constrained Mean-Variance Portfolio Optimisation Problem (CCMVPOP)~\cite{card_portf_master} considered in this study entails selecting assets that maximise returns while minimising the associated risks. Typically, multi-objective problems are converted to single objective problems before the Ising machines are applied to them. For example, the $\epsilon$-constraint approach was used in the Quantum Annealer to solve a portfolio optimisation problem~\cite{phillipson2021portfolio}. Scalarisation has also been used when solving multi-objective QUBO in previous work \cite{zhou2018multi,Ayodele2022Multi}. One of the main challenges to using scalarisation is how to define a set of weights resulting in a diverse set of solutions on the Pareto front (PF). A common approach is to generate weights uniformly using, for example, the simplex lattice design \cite{zhou2018multi}. However, a uniform choice of weights does not necessarily translate to a diverse set of Pareto-optimal solutions~\cite{liefooghe2015experiments,zhou2018multi}. Previous studies have therefore also considered iterative method which uses a dichotomic procedure to derive new weights perpendicular to two solutions that have the largest distance between them \cite{dubois2011improving,liefooghe2015experiments}. In this study, we propose a method for deriving scalarisation weights which targets less explored regions of the PF. The proposed method utilises the weights used during previous scalarisations in addition to the relative position of the corresponding solutions in the PF and relies less on the weights and fitness being perfectly correlated. 

The following section presents the problem description of the CCMVPOP. Methods of generating scalarisation weights used in this study are described in Section \ref{sec:sbda}. Results and conclusions are presented in Sections \ref{sec:res} and \ref{sec:con}.

\section{Cardinality Constrained Mean-Variance Portfolio Optimisation Problem}
\label{sec:probform}
%\vspace{-0.4cm}
Portfolio Optimisation entails selecting assets that maximise returns while minimising the associated risks. In the CCMVPOP~\cite{card_portf_master}, cardinality constraints on the  number of asset types to be considered are imposed. Given the number of asset types to consider ($n$), the fixed number of assets a portfolio must contain ($K$), the expected return of asset $i$ ($\mu_i$) and the covariance between assets $i$ and $j$ ($\sigma_{i,j}$), the minimum ($\epsilon_i$) and maximum ($\delta_i$) proportion of a chosen asset $i$, we aim to find the proportion of each asset $i$ to hold ($w_i\in[0,1]$). Binary variables $z_i$ are used to indicate whether an asset $i$ is selected or not. The CCMVPOP is formally defined as follows.
\begin{align}
\label{eqn:objective}
%\text{minimize}   &\quad \lambda \sum_{i=1}^{n}\sum_{j=1}^{n}{w_iw_j\sigma_{i,j}} - (1-\lambda) \sum_{i=1}^{n}{w_i\mu_i} 
\text{minimise} & \quad \lambda_1\left (   \sum_{i=1}^{n}\sum_{j=1}^{n}{w_iw_j\sigma_{i,j}} \right )+  \lambda_2 \left (  -\sum_{i=1}^{n}{w_i\mu_i} \right )\\
%\end{align}
%
%\begin{align}
%\text{subject to} &  \nonumber\\
\label{eqn:proportion}
 \text{subject to} & \quad \sum_{i=1}^{n}{w_i}=1, \quad \sum_{i=1}^{n}{z_i}=K\\
% \end{align}
%
%\begin{align}
\label{eqn:min_max_prop}
& \epsilon_iz_i\leq w_i \leq \delta_iz_i, \quad z_i \in \{0,1\},\quad i=1,\ldots,n
\end{align}

The first objective is the first term in Eq.~\eqref{eqn:objective} and minimises the risk (sum of covariance between all pairs $i$,$j$ of chosen assets) of the chosen assets of the portfolio. The second objective is the second term in Eq.~\eqref{eqn:objective} and maximises returns (sum of expected return of each asset $i$) of chosen assets. A negative sign is appended to the second objective to convert it to a minimisation problem. $\lambda = (\lambda_1, \lambda_2)$ is a set of scalarisation weights. The cardinality constraint (Eq. \ref{eqn:proportion}) forces the number of chosen assets to be equal to $K$, and Eq. \eqref{eqn:min_max_prop} ensures the proportion of a chosen asset $w_i$ to be within given bounds. The QUBO formulation of the CCMVPOP ($K=10$, $\epsilon_i=0.01$ $\delta_i=1$) used is based on the binary representation presented in \cite{Parizy2022Cardinality}.

\section{Scalarisation Methods}
\label{sec:sbda}
%\vspace{-0.4cm}
In \cite{Ayodele2022Multi}, a scalarisation framework, Scalarisation Based DA (SB-DA), was proposed for obtaining multiple non-dominated solutions for the bi-objective quadratic assignment problem formulated as QUBO. A CPU implementation of the 1$^\text{st}$ generation DA algorithm \cite{aramon2019physics} was used in that study. However, in this study, we use the $3^\text{rd}$ generation DA \cite{da3} which is designed to be faster and more efficient than previous generations of the DA, it also benefits from hardware speedup \cite{da3}. For simplicity, we use DA to refer to $3^\text{rd}$ generation DA in the rest of this work.

\begin{algorithm}[t]
\caption{SB-DA Algorithm}\label{alg:sbda}
\begin{algorithmic}[1]
\Require $B$, $D$, $G$, $k$, $time$,  $n\_{top}, s\_type$
\State $\Lambda \assign \left \{ (0,1), (1,0) \right \} $ 
\sIfElse{$s\_type$ in \{\textit{random}, \textit{uniform}\}}{Mode $\assign$ static}{Mode $\assign$ iterative}
\sIf{$s\_type$ is \textit{random}}{add $k-2$ sets of random weights to $\Lambda$}
\sIf{$s\_type$ is \textit{uniform}}{ $\Lambda \assign$ SLD($H=k$, $m=2$)}\label{sld}
\sIfElse{Mode is static}{$A \assign$ execute SB-DA$s$}{$A \assign$ execute SB-DA$i$}
\State \Return all non-dominated solutions from archive $A$ 
\end{algorithmic}
\end{algorithm}

\begin{algorithm}[htb]
\caption{SB-DA$i$}\label{alg:sbdait}
\begin{algorithmic}[1]
\Require $B$, $D$, $G$, $k$, $time$,  $n\_{top}$, $\Lambda$
\State $A \assign \emptyset$, $W \assign \left \{ \right \}$   \Comment{Initialise archive and mapping between weights and solutions}
\ForEach{$i \in \left \{ 1,\dots ,k \right \}$ }
%\sIfElse{$i \leq 2$}{ $\lambda  = \left ( \lambda_1, \lambda_2 \right ) \assign \Lambda_i$ \quad $R, S \assign B, D$}{$R, S \assign rescale(B, D)$}, $\lambda \assign \emptyset$, $max\_d \assign 0$  \label{ln:rescale2} \Comment{Initialise weights and maximum distance}
\If{$i \leq 2$}
\State  $\lambda  = \left ( \lambda_1, \lambda_2 \right ) \assign \Lambda_i$ \quad $R, S \assign B, D$
\Else
\State $R, S \assign rescale(B, D)$ \label{ln:rescale2}
%\EndIf
\State  $\lambda \assign \emptyset$, $max\_d \assign 0$ \Comment{Initialise weights and maximum distance}
\For{$j \in [1, i-2]$}
\State $d \assign Distance(W_j, W_{j+1})$  \Comment{Manhattan distance} \label{ln:manhattah}
\State $[(\lambda^{sol1}_1, \lambda^{sol1}_2), sol1], [(\lambda^{sol2}_1, \lambda^{sol2}_2), sol2] \assign$ $W_j$, $W_{j+1}$
\State $\lambda\_temp \assign \left (avg(\lambda^{sol1}_1,\lambda^{sol2}_1), avg(\lambda^{sol1}_2,\lambda^{sol2}_2) \right )$  \label{ln:avglambda}
\sIf{$(d > max\_d)$ and ($\lambda\_temp \notin  W$)}{$\lambda \assign \lambda\_temp$, $max\_d \assign d$ } \label{ln:condition}
\EndFor
\sIf{$\lambda == \emptyset$}{$\lambda \assign$ Random weights} \Comment{each set of weights sums to 1}
\EndIf
\State $Q \assign (\lambda_1 \cdot  R +  \lambda_2 \cdot S) + \alpha \cdot G$ \label{q2}
\State $Y \assign$ ExecuteDA($Q$, $n\_{top}$, $time\_limit= \frac{T}{k}$), \quad add all solutions in $Y$ to $A$ \label{executeda2}
%\State add all solutions in $Y$ to $A$
\State $W_i \assign [\lambda, Y_0]$ \Comment{save weight and best solution in $Y$}
\EndFor
\State \Return $A$ 
\end{algorithmic}
\end{algorithm}

\begin{algorithm}[t]
\caption{SB-DA$s$}\label{alg:sbdaru}
\begin{algorithmic}[1]
\Require $B$, $D$, $G$, $k$, $time$,  $n\_{top}$, $\Lambda$ 
\State $A \assign \emptyset$\Comment{Initialise archive}
\ForEach{$i \in \left \{ 1,\dots ,k \right \}$ }
\State  $\pmb{\lambda}  = \left ( \lambda_1, \lambda_2 \right ) \assign \Lambda_i$
\sIfElse{$i > 2$}{$R, S \assign rescale(B, D)$ }{$R, S \assign B, D$}\label{ln:rescale1}
\State $Q \assign (\lambda_1 \cdot  R +  \lambda_2 \cdot S) + \alpha \cdot G$ \label{q1}
\State $Y \assign$ ExecuteDA($Q$, $n\_{top}$, $time\_limit= \frac{T}{k}$), add all solutions in $Y$ to $A$ \label{executeda1}
%\State $Y \assign$ ExecuteDA($R$, $S$, $G$, $\lambda$, $n\_{top}$, $time\_limit = \frac{T}{k}$) \label{executeda1}
%\State add all solutions in $Y$ to $A$
\EndFor
\State \Return $A$ 
\end{algorithmic}
\end{algorithm}

We propose two extensions of the SB-DA, which we call SB-DA$s$ and SB-DA$i$ (Alg. \ref{alg:sbda}). Parameters $B$, $D$ and $G$ are QUBO matrices representing the first objective, second objective and constraint functions, respectively. The number of scalarisation weights is denoted by $k$ and $time$ is the total time allowed for all DA executions. To allow more solutions to be considered for non-dominance, $n\_{top}$ is a parameter used to define the number of top solutions (solutions with the lowest energies) to be returned during each DA execution. In this study, we compared three methods of deriving scalarisation weights, $s\_type$ set to \textit{random}, \textit{uniform} or \textit{iterative}. Where $s\_type$ is set to \textit{random}, $k$ sets of randomly generated weights are pre-computed. For each set of weights $\lambda = (\lambda_1, \lambda_2)$, $\lambda_1$ is a random value between range [0, 1] while $\lambda_2 = 1-\lambda_1 $. For $s\_type$ set to \textit{uniform} method, $k$ sets of evenly distributed weights are pre-computed. In this study, we use the the Simplex Lattice Design (SLD) (Line \ref{sld} of Alg. \ref{alg:sbda}) to generate evenly distributed weights. Where $s\_type$ is \textit{iterative}, weights are derived with the aim of finding solutions that fall within the less crowded region of the Pareto front. To achieve this aim, a mapping between each set of weights and the best solution found by the DA using such set of weights are stored as $W$. $W$ is sorted in ascending order of $\lambda_1$. For any two adjacent solutions in $W$, the manhatthan distance between the solutions are recorded. Solutions $sol1$ and $sol2$ that correspond to the largest gap in the Pareto front are saved. A set of scalarisation weights used to derive $sol1$ and $sol2$ are $\lambda^{sol1} =(\lambda^{sol1}_1, \lambda^{sol1}_2)$ and $\lambda^{sol2} =(\lambda^{sol2}_1, \lambda^{sol2}_2)$ respectively. An average of $\lambda^{sol1}$ and $\lambda^{sol2}$ becomes the scalarisation weights used in the new iteration  (Lines \ref{ln:manhattah}-\ref{ln:avglambda}). The new set of weights is however only used if this has not been used in previous iterations (Line \ref{ln:condition}). If there is no unique set of weights that can be derived using this procedure, randomly generated weights are used.

SB-DA can be executed in one of two modes. It is executed in static mode (SB-DA$s$: Alg. \ref{alg:sbdaru}), if $s\_type$ is \textit{random} or \textit{uniform} and in iterative mode (SB-DA$i$: Alg. \ref{alg:sbdait}) if $s\_type$ is set to \textit{iterative}. In SB-DA$s$, scalarisation weights are pre-computed while in SB-DA$i$, a set of scalarisation weights at a given iteration is influenced by the set of scalarisation weights used in previous iterations. An iteration of SB-DA refers to a run of the DA with a given set of scalarisation weights. In both modes of the SB-DA, each objective is optimised independently ($\lambda = (0,1)$ and $\lambda = (1,0)$) before other weights are used. This is because QUBO matrices $B$ and $D$ are rescaled using information about the Lower Bound (LB) and Upper Bound (UB). These bounds are achieved by minimising each objective independently. LB of $B$ (or $D$) is derived by minimising $B$ (or $D$) independently. Conversely, UB of $B$ (or $D$) is derived by minimising $D$ (or $B$) independently. The LB or UB are updated if smaller or larger energies are found for any individual objective at any iteration of the SB-DA. In Line \ref{ln:rescale1} of Alg. \ref{alg:sbdaru} and Line \ref{ln:rescale2} of Alg. \ref{alg:sbdait}, $rescale(B,D)$ is computed such that $R =  \max_{1\leq i \leq k}$(UB$_i$)/(UB$_1$-LB$_1$)$\cdot B$ and $S = \max_{1\leq i \leq k}$(UB$_i$)/(UB$_2$-LB$_2$)$\cdot D$. This is done to reduce bias towards any of the objectives, allowing the algorithm to control the bias using the scalarisation weights only. QUBO matrix $Q$ (Alg. \ref{alg:sbdaru}: Line \ref{q1}, Alg. \ref{alg:sbdait}:Line \ref{q2}) is an aggregate of QUBO matrices representing the objectives and constraint, penalty weight ($\alpha$) is set using Maximum change in Objective function divided by Minimum Constraint function of infeasible solutions (MOMC) originally proposed in \cite{ayodele2022penalty}. ExecuteDA (Alg. \ref{alg:sbdaru}: Line \ref{executeda1}, Alg. \ref{alg:sbdait}: Line \ref{executeda2}) runs DA on $Q$ for $time\_limit$ seconds, returning the non-dominated solutions found amongst the best $n\_top$ solutions. Note that SB-DA proposed in \cite{Ayodele2022Multi} is equivalent to the proposed extended SB-DA when executed with $s\_type$ set to uniform and $n\_{top} = 1$.

\section{Results}
\label{sec:res}
%\vspace{-1em}
%\vspace{-1.3cm}
\begin{table}[b]
\centering
\caption{Average and standard deviation number of non-dominated solutions and hypervolume across 20 runs of the DA (stopping criteria: $0.05n$, Number of weights: 10). Hypervolume values have been divided by $10^{23}$. Reference points used for computing the hypervolume values are the sum of all positive QUBO coefficients in $B$ and $D$
$\left (  \sum_{i=1}^{m}\sum_{i=1}^{m}\max\{0, B_{ij}\}, \sum_{i=1}^{m}\sum_{i=1}^{m}\max\{0, D_{ij}\}\right )$.
  % ($(\left (  \sum_{i=1}^{m}\sum_{i=1}^{m}\max(0, B_{ij}), \sum_{i=1}^{m}\sum_{i=1}^{m}\max(0, D_{ij}))\right )$).
  Best values presented in bold (student t-test used for test of significance). \label{tb:port}}
\resizebox{1\columnwidth}{!}{
\begin{tabular}{@{}cccc|@{}}
\toprule
\begin{tabular}[c]{@{}c@{}}Problem   \\ Instance (n) \end{tabular} & \begin{tabular}[c]{@{}c@{}}Scalarisation\\ Method\end{tabular} & \begin{tabular}[c]{@{}c@{}}Mean \textpm Stdev  \\ Non-dominated \\ solutions \end{tabular} &  \begin{tabular}[c]{@{}c@{}}Mean \textpm Stdev \\ Hypervolume\end{tabular} \\ \midrule
\multirow{3}{*}{\begin{tabular}[c]{@{}c@{}} Port1 \\ (248) \end{tabular} } & Random & 324 \textpm 50 & 37.06 \textpm  14.38  \\
 & Uniform &  \textbf{354}\textpm 39 & 28.71  \textpm  9.50   \\
 & Iterative & 250\textpm 25 & \textbf{60.37} \textpm  0.22  \\
 \midrule
\multirow{3}{*}{\begin{tabular}[c]{@{}c@{}} Port2 \\ (680) \end{tabular} } &  Random & 514 \textpm 82 & 159.92 \textpm  7.79 \\
 & Uniform & \textbf{676} \textpm 61 & 155.38 \textpm  0.86  \\
 & Iterative & 275 \textpm 50 & \textbf{164.44} \textpm  0.92  \\
\bottomrule
\end{tabular}
 \begin{tabular}{@{}cccc@{}}
\toprule
\begin{tabular}[c]{@{}c@{}}Problem   \\ Instance (n) \end{tabular} & \begin{tabular}[c]{@{}c@{}}Scalarisation\\ Method\end{tabular} & \begin{tabular}[c]{@{}c@{}}Mean \textpm Stdev  \\ Non-dominated \\ solutions \end{tabular} &  \begin{tabular}[c]{@{}c@{}}Mean \textpm Stdev \\ Hypervolume \end{tabular} \\ \midrule
\multirow{3}{*}{\begin{tabular}[c]{@{}c@{}} Port3 \\ (712) \end{tabular} } & Random & 628 \textpm  132 &  121.64 \textpm  25.30  \\
 & Uniform & \textbf{866}\textpm 51 &94.27 \textpm  1.90 \\
 & Iterative & 417\textpm 71 &\textbf{136.46} \textpm  1.45 \\
 \midrule
\multirow{3}{*}{\begin{tabular}[c]{@{}c@{}} Port4 \\ (784) \end{tabular} }  & Random & 552 \textpm 112 &117.79 \textpm  25.35  \\
 & Uniform & \textbf{798}\textpm 49 & 116.73 \textpm  1.23 \\
 & Iterative & 398 \textpm 49 & \textbf{136.42} \textpm  2.25 \\
\bottomrule
\end{tabular}
}
\end{table}

To generate the results presented in this section, default parameters of the DA are used. Table \ref{tb:port} shows that the \textit{uniform} method consistently found the highest number of non-dominated solutions while the \textit{iterative} method consistently found the lowest number of non-dominated solutions across all problem instances. However, using the proposed \textit{iterative} method consistently led to the highest hypervolume. Higher hypervolume values were reached because the iterative method was able to find weights that allowed the algorithm to focus on harder and more extreme regions of the search space.

\begin{figure}[th]
  \centering%
   \includegraphics[width=0.4\columnwidth]{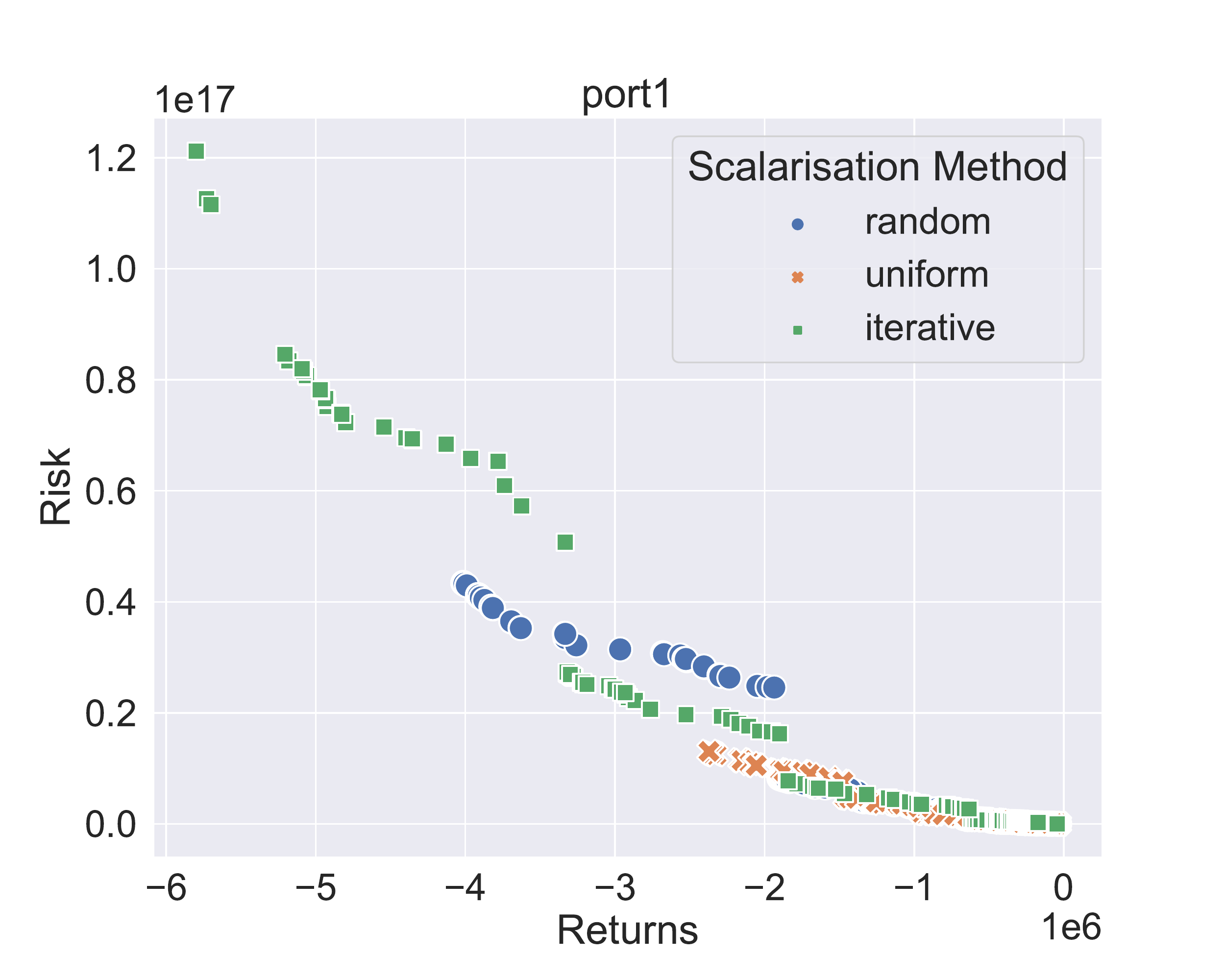}
     \includegraphics[width=0.4\columnwidth]{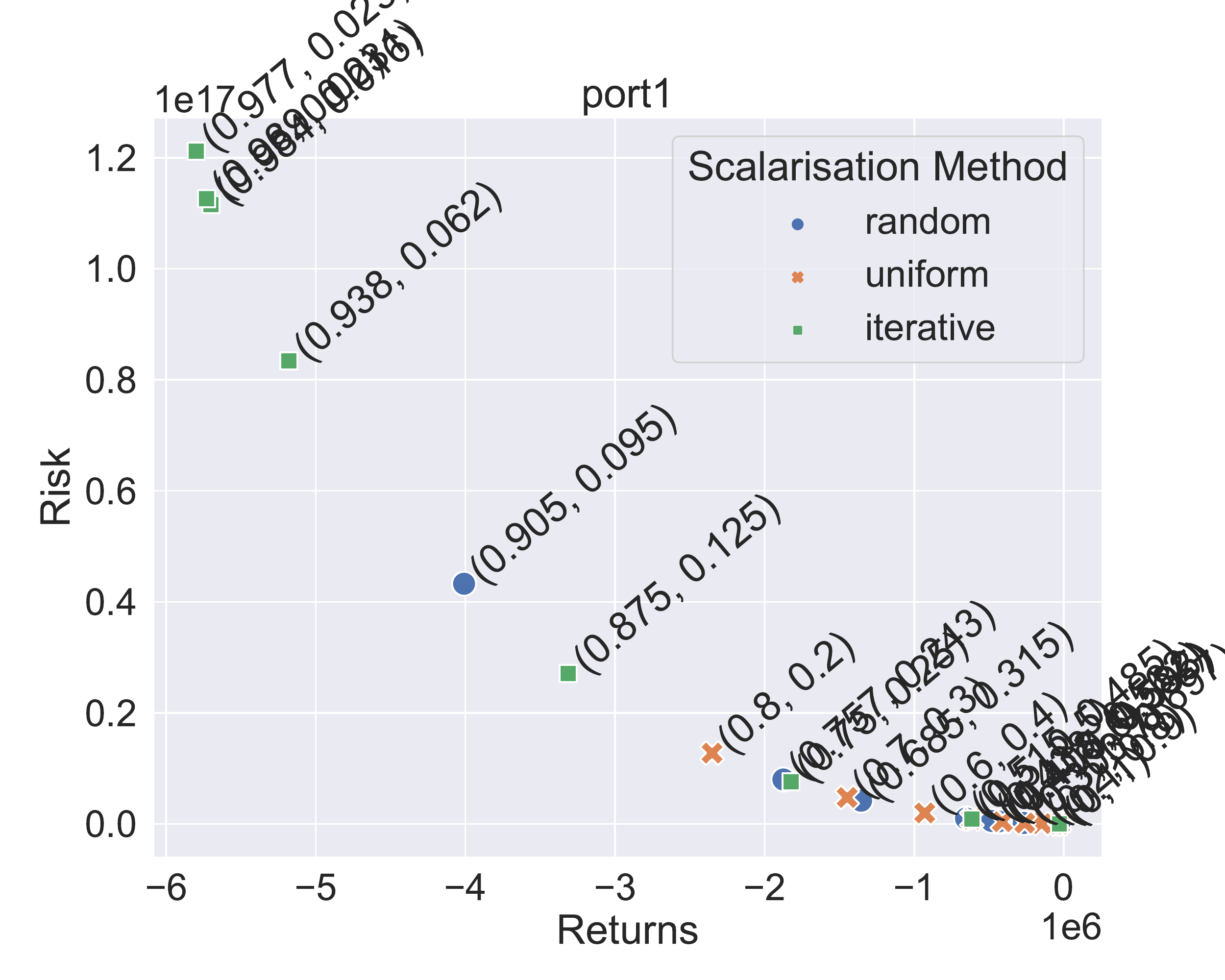}
  \caption{Comparing Scalarisation Methods (stopping criteria: $0.05n$ seconds, Number of weights: 10). plots show the best solution(s) produced by each method (left plot:$n\_top$ =1 and annotated by weights that resulted in the corresponding best solution, right plot:$n\_top$ =1000)}

   \label{fig:scatter1}
\end{figure}

\begin{figure}[th]
  \centering%
        \includegraphics[width=0.4\columnwidth]{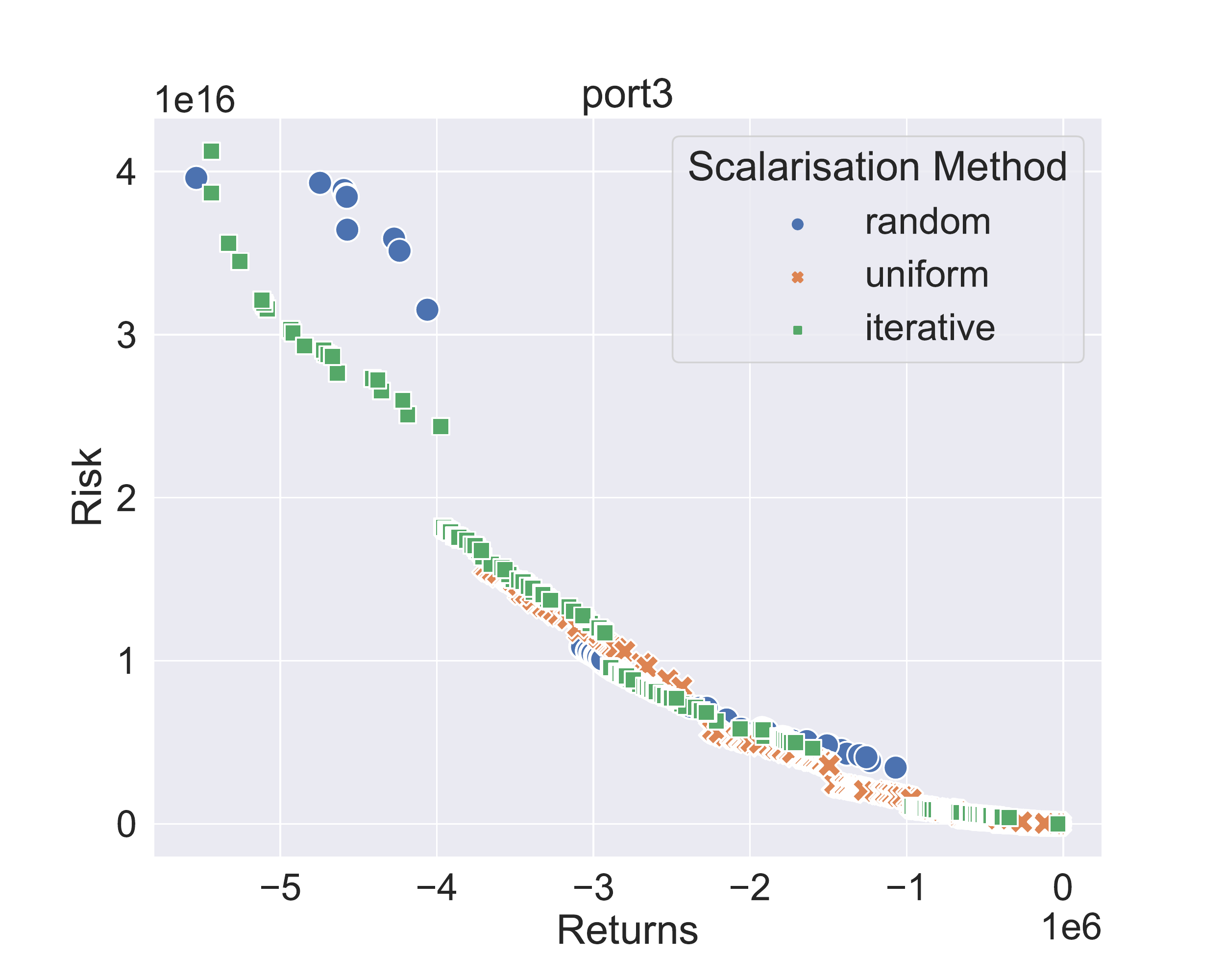}
     \includegraphics[width=0.4\columnwidth]{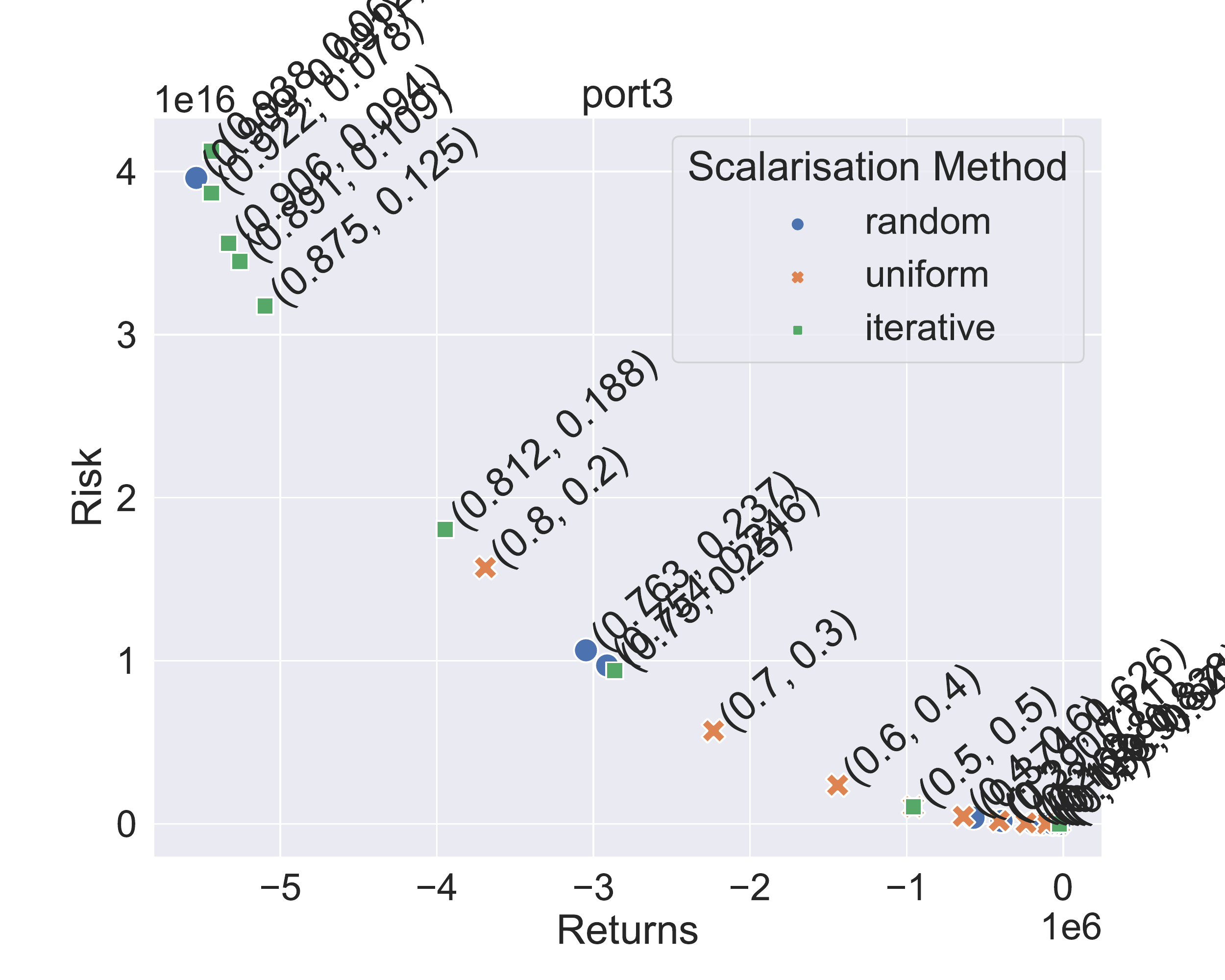}
        \includegraphics[width=0.4\columnwidth]{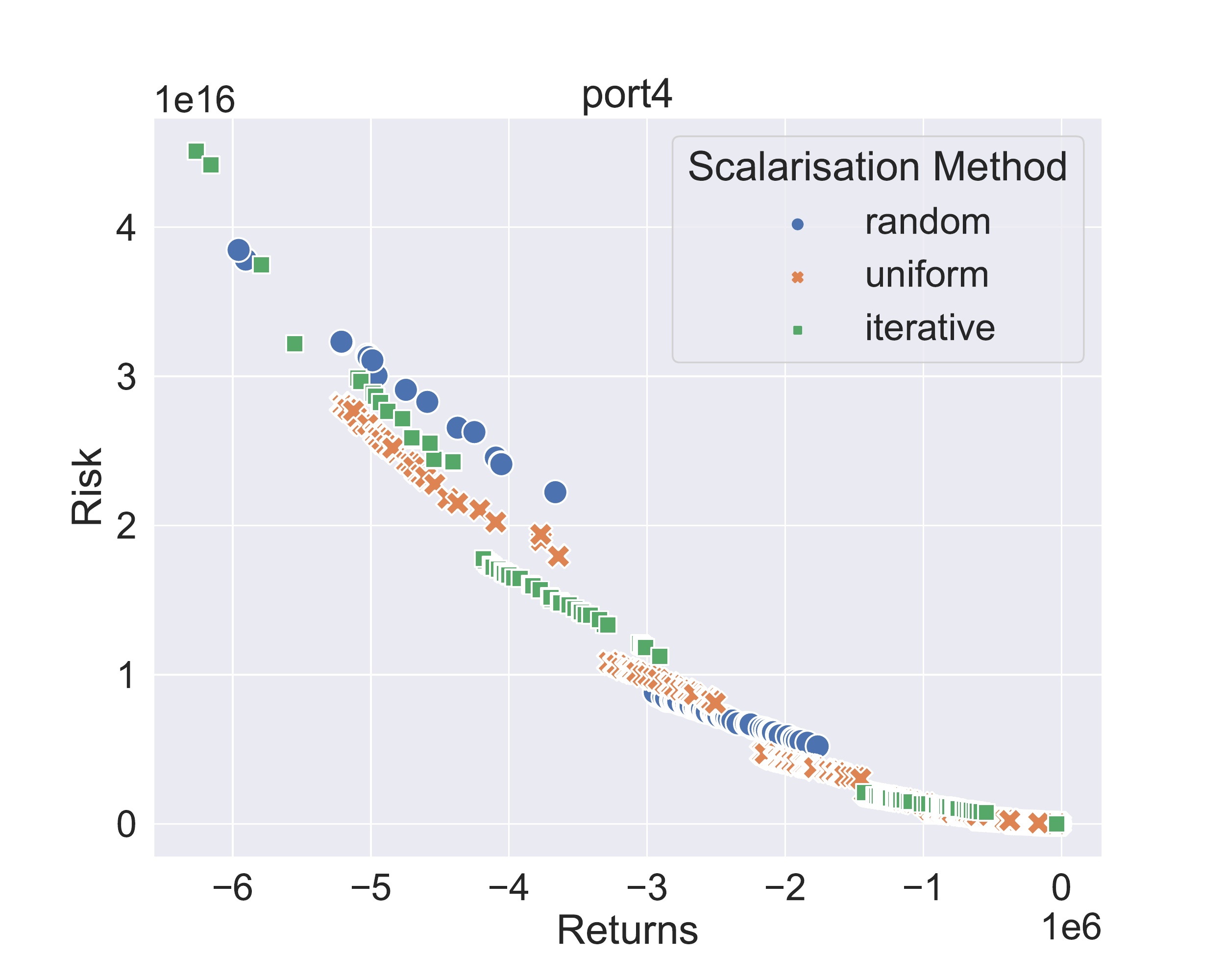}
     \includegraphics[width=0.4\columnwidth]{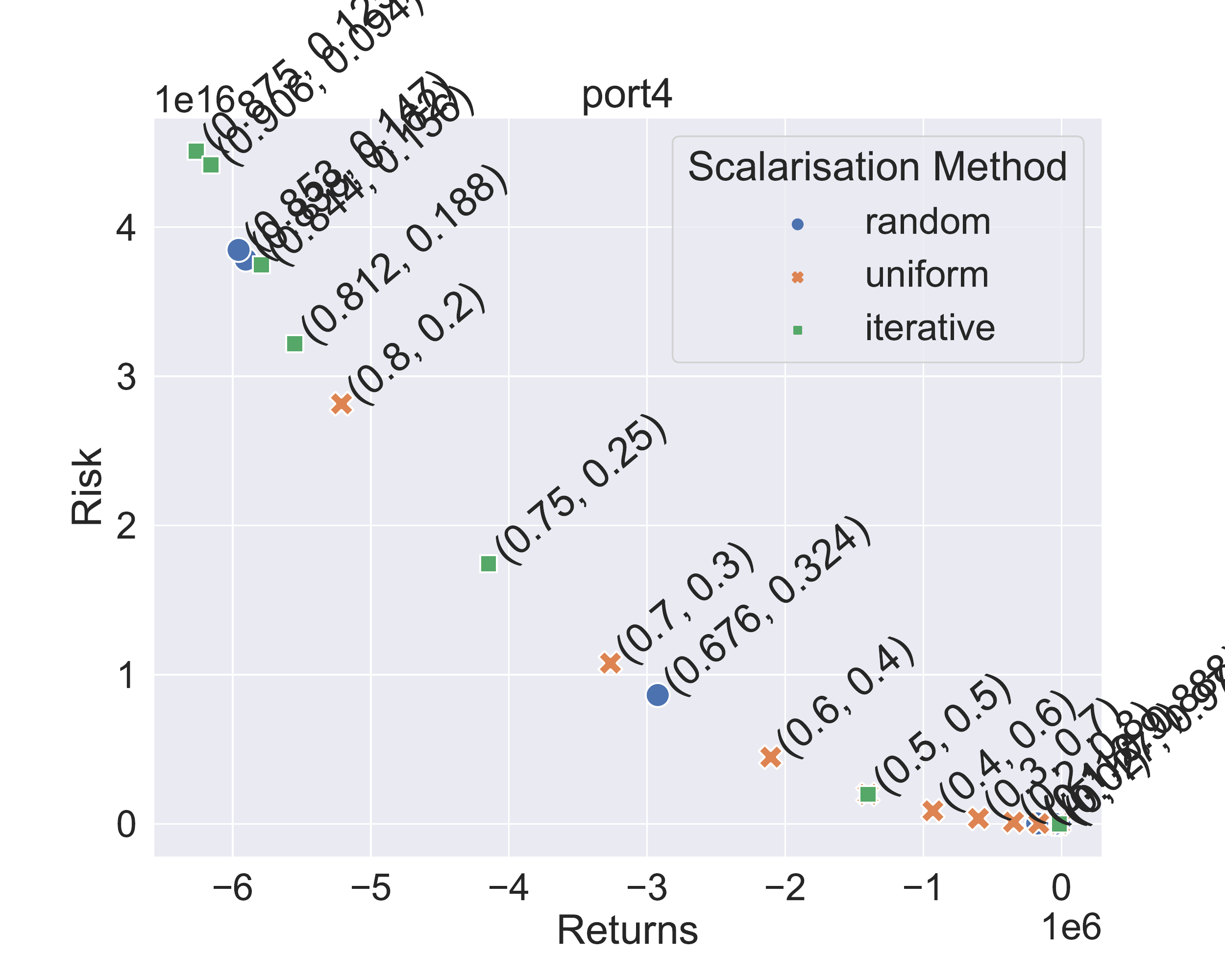}
  \caption{Comparing Scalarisation Methods (stopping criteria: $0.05n$ seconds, Number of weights: 10). plots show the best solution(s) produced by each method (left plot:$n\_top$ =1 and annotated by weights that resulted in the corresponding best solution, right plot:$n\_top$ =1000)}

   \label{fig:scatter2}
\end{figure}

\begin{figure}[thb]
  \centering%
   \includegraphics[width=0.49\columnwidth]{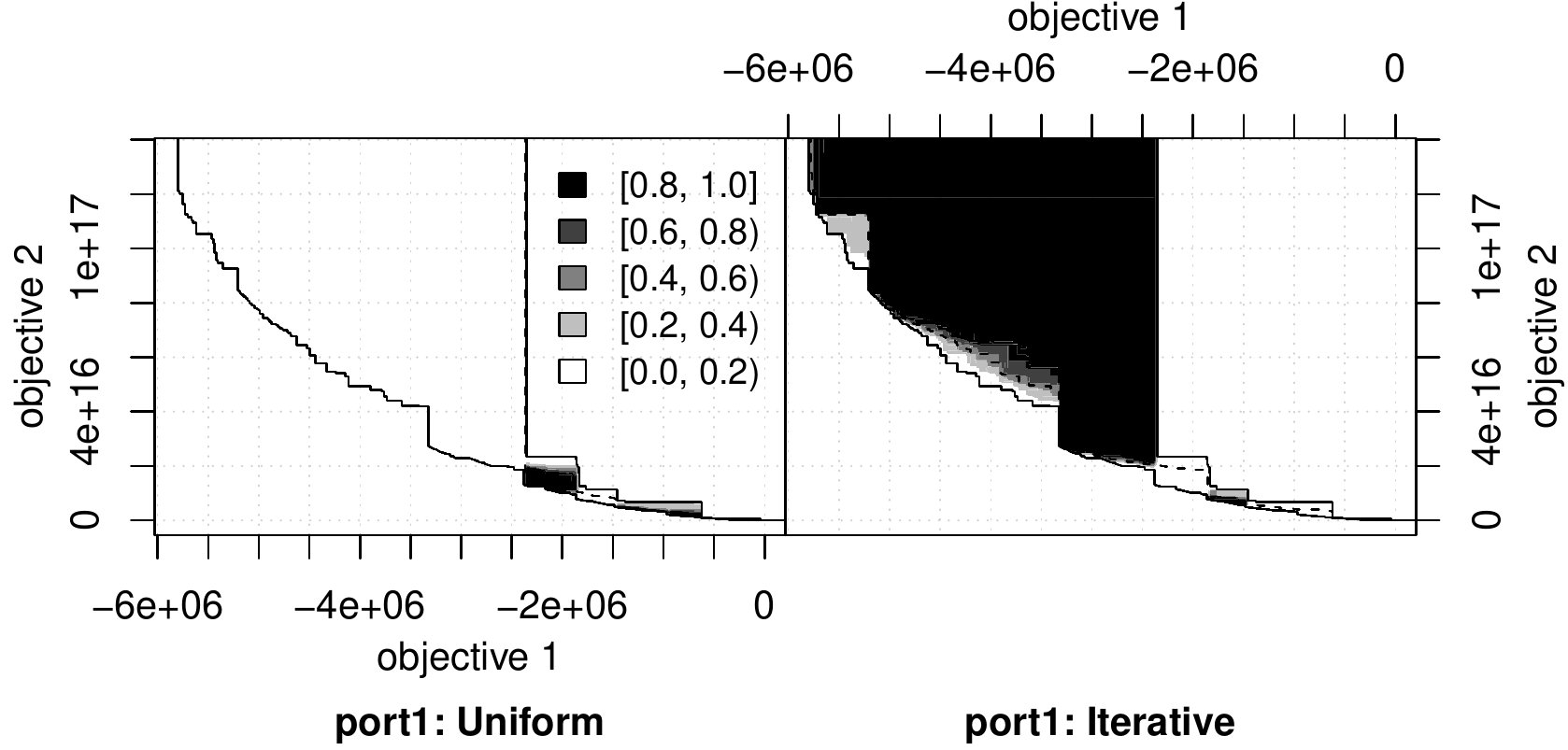}
\includegraphics[width=0.49\columnwidth]{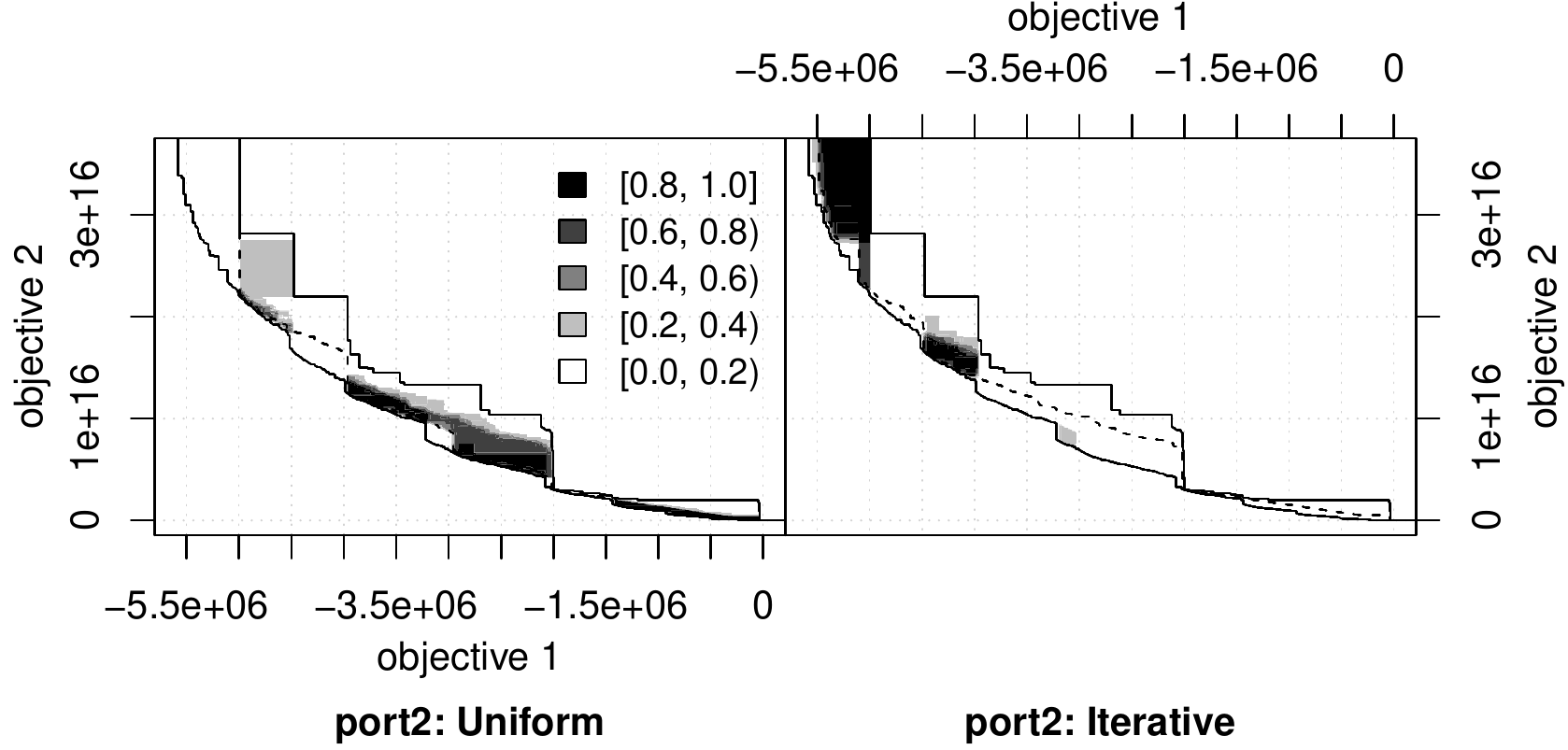}
  \caption{Comparing Scalarisation Methods (stopping criteria: $0.05n$ seconds, Number of weights: 10). Empirical Attainment Surface based on 20 runs ($n\_top$ =1000). Objective values for returns (maximisation) are shown as negative (minimisation). Objectives 1 and 2 and respectively used to denote \textit{returns} and \textit{risk}}
   \label{fig:eaf1}
\end{figure}

\begin{figure}[thb]
  \centering%
\includegraphics[width=0.49\columnwidth]{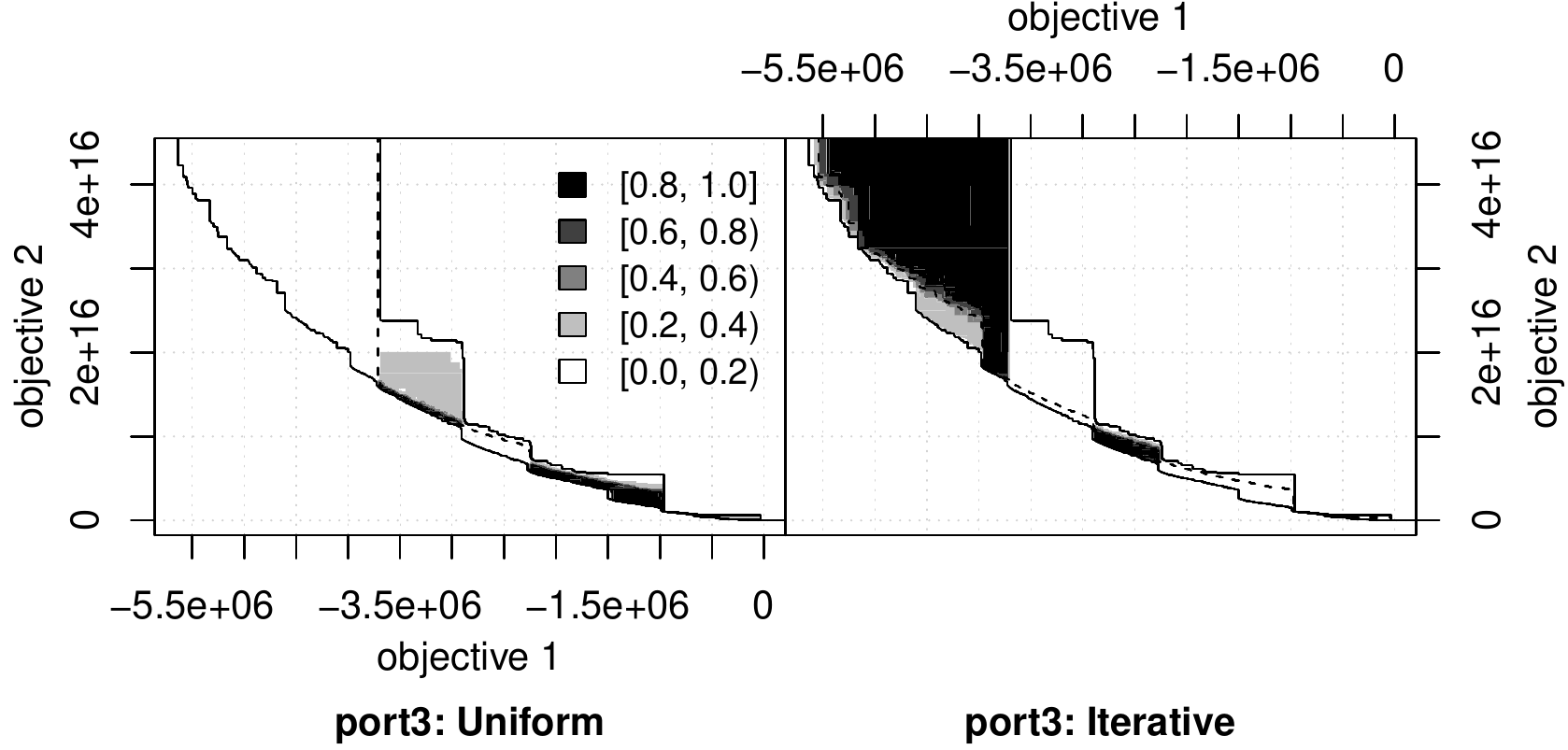}
\includegraphics[width=0.49\columnwidth]{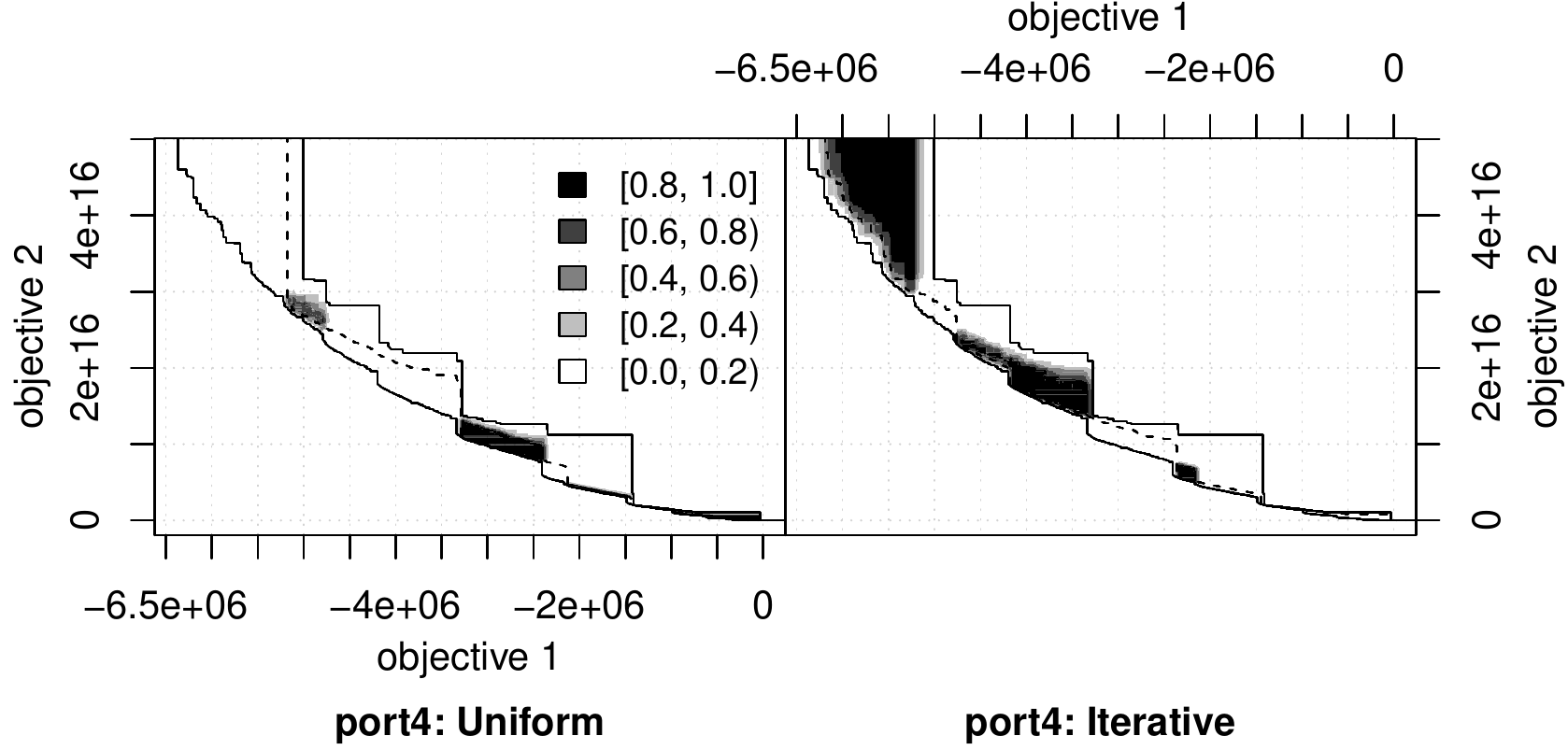}
  \caption{Comparing Scalarisation Methods (stopping criteria: $0.05n$ seconds, Number of weights: 10). Empirical Attainment Surface based on 20 runs ($n\_top$ =1000). Objective values for returns (maximisation) are shown as negative (minimisation). Objectives 1 and 2 and respectively used to denote \textit{returns} and \textit{risk}}
   \label{fig:eaf2}
\end{figure}

 To illustrate the difference between the bahaviour of the methods, we show an example of a single run on the CCMVPOP instances in Figures \ref{fig:scatter1} and \ref{fig:scatter2}. This shows that the \textit{iterative} method allowed the algorithm to concentrate on regions of the Pareto front that are harder to reach. To show that this behaviour is consistent across multiple runs. We also present comparison of \textit{iterative} and \textit{uniform} using the empirical attainment surface \citep{lopez2010exploratory} in Figures \ref{fig:eaf1} and \ref{fig:eaf2} (darker regions show parts where one method is better than the other). This is a visualisation of the Empirical Attainment Function (EAF). The EAF of an algorithm is the probability, estimated from multiple runs, that the non-dominated set produced by a single run of the algorithm dominates a particular point in the objective space. The visualisation of the EAF~\citep{Grunert01} has been shown as a suitable graphical interpretation of the quality of the outcomes returned by local search methods. The visualisation of the differences between the EAFs of two alternative algorithms indicates how much better one method is compared to another in a particular region of the objective space~\citep{LopPaqStu09emaa}. The EAF visualisations were done using the \texttt{eaf} \textsf{R} package.\footnote{\url{http://lopez-ibanez.eu/eaftools}}. In Figure \ref{fig:eaf1} and \ref{fig:eaf2}, we particularly see more darker regions (indicating better performance) using the \textit{iterative} method on 'Port1' and 'Port3' instances compared to the \textit{uniform} method.

\section{Conclusions}
\label{sec:con}
%\vspace{-0.3cm}
 This study compared three simple methods of generating scalarisation weights within the context of bi-objective QUBO solving. The methods were applied to QUBO formulation of the CCMVPOP. We show that considering more than one best solution during each scalarisation can lead to finding more non-dominated solutions. We also show that for this problem, higher hypervolume can be reached by using adaptive methods of generating scalarisation weights when compared to random or evenly distributed weights.

Some areas of future work include, 1. comparing approach on different problems. 2. comparing approach on problems with more than two objectives.

% \begin{acks}
% M.\@ L\'opez-Ib\'a\~nez is a ``Beatriz Galindo'' Senior Distinguished Researcher (BEAGAL 18/00053) funded by the Spanish Ministry of Science and Innovation (MICINN).
% \end{acks}

\bibliographystyle{unsrtnat}
\bibliography{bib/abbrev,bib/journals,bib/authors,bib/biblio,bib/crossref,References}

%\appendix
%\section{}

\end{document}